\documentclass[10pt,twocolumn,letterpaper]{article}

\usepackage{iccv}
\usepackage{times}
\usepackage{epsfig}
\usepackage{graphicx}
\usepackage{amsmath}
\usepackage{amssymb}

\usepackage{acronym}
\usepackage{enumitem}
\usepackage{algorithm}
\usepackage[noend]{algpseudocode}
\usepackage{xspace}
\usepackage{balance}
\usepackage{setspace}
\usepackage{color}
\usepackage[table]{xcolor}
\usepackage{colortbl}
\usepackage{xpatch}
\usepackage{tabularx}
\definecolor{mygray}{gray}{.92}

\DeclareMathOperator*{\argmax}{arg\,max}
\DeclareMathOperator*{\argmin}{arg\,min}
\makeatletter 
\newcommand{\multiline}[1]{%
  \begin{tabularx}{\dimexpr\linewidth-\ALG@thistlm}[t]{@{}X@{}}
    #1
  \end{tabularx}
}
\makeatother

\usepackage[pagebackref=true,breaklinks=true,letterpaper=true,colorlinks,bookmarks=false]{hyperref}

\iccvfinalcopy 

\def\iccvPaperID{****} 

\ificcvfinal\fi

\begin{document}

\makeatletter
\renewcommand{\paragraph}{%
  \@startsection{paragraph}{4}%
  {\z@}{0ex \@plus 0ex \@minus 0ex}{-1em}%
  {\hskip\parindent\normalfont\normalsize\bfseries}%
}
\makeatother

\acrodef{mcmc}[MCMC]{Markov chain Monte Carlo}
\acrodef{hoi}[HOI]{human-object interaction}
\acrodef{mrf}[MRF]{Markov random field}
\acrodef{map}[MAP]{maximum a posteriori}
\acrodef{iou}[IoU]{intersection-over-union}
\acrodef{mlp}[MLP]{multi-layer perceptron}

\title{\vspace{-12pt}Holistic$^{++}$ Scene Understanding:\\Single-view 3D Holistic Scene Parsing and Human Pose Estimation\\with Human-Object Interaction and Physical Commonsense\vspace{-12pt}}

\author{Yixin Chen$^{\star 1}$, Siyuan Huang$^{\star 1}$, Tao Yuan$^1$, Siyuan Qi$^{1,2}$, Yixin Zhu$^{1,2}$, and Song-Chun Zhu$^{1,2}$\\
{\small $^{\star}$ Equal Contributors}
\vspace{3pt}\\
$^{1}$ University of California, Los Angeles (UCLA)\\
$^{2}$ International Center for AI and Robot Autonomy (CARA)\\
{\tt\small \{ethanchen,huangsiyuan,taoyuan,syqi,yixin.zhu\}@ucla.edu,sczhu@stat.ucla.edu}
}

\maketitle
\ificcvfinal\thispagestyle{empty}\fi

\begin{abstract}
We propose a new 3D holistic$^{++}$ scene understanding problem, which jointly tackles two tasks from a single-view image: (i) holistic scene parsing and reconstruction---3D estimations of object bounding boxes, camera pose, and room layout, and (ii) 3D human pose estimation. The intuition behind is to leverage the coupled nature of these two tasks to improve the granularity and performance of scene understanding. We propose to exploit two critical and essential connections between these two tasks: (i) \acf{hoi} to model the fine-grained relations between agents and objects in the scene, and (ii) physical commonsense to model the physical plausibility of the reconstructed scene. The optimal configuration of the 3D scene, represented by a parse graph, is inferred using \acf{mcmc}, which efficiently traverses through the non-differentiable joint solution space. Experimental results demonstrate that the proposed algorithm significantly improves the performance of the two tasks on three datasets, showing an improved generalization ability.
\end{abstract}

\section{Introduction}

\begin{figure}[t!]
    \centering
    \includegraphics[width=\linewidth,trim={0.5cm 0.8cm 0.3cm 1.0cm},clip]{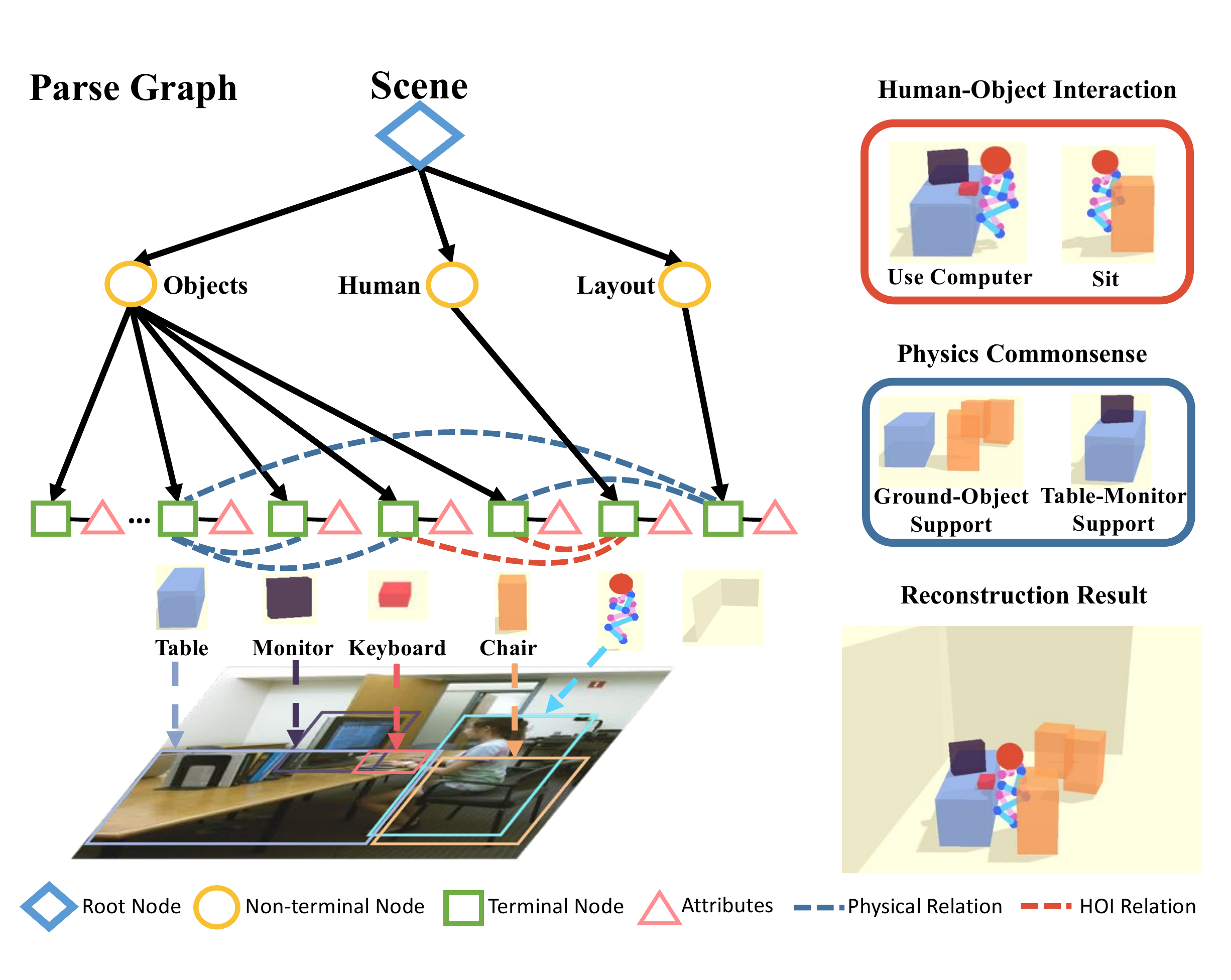}
    \caption{\textbf{holistic$^{++}$ scene understanding} task requires to jointly recover a parse graph that represents the scene, including human poses, objects, camera pose, and room layout, all in 3D. Reasoning \acf{hoi} helps reconstruct the detailed spatial relations between humans and objects. Physical commonsense (\eg, physical property, plausibility, and stability) further refines relations and improves predictions.}
    \label{fig:pg}
\end{figure}

Humans, even young infants, are adept at perceiving and understanding complex indoor scenes. Such an incredible vision system not only relies on the data-driven pattern recognition but also roots from the visual reasoning system, known as the core knowledge~\cite{spelke2007core}, that facilitates the 3D holistic scene understanding tasks. Consider a typical indoor scene shown in \autoref{fig:pg} where a person sits in an office. We can effortlessly extract rich knowledge from the static scene, including 3D room layout, 3D position of all the objects and agents, and correct \acf{hoi} relations in a physically plausible manner. In fact, psychology studies have established that even infants employ at least two constraints---\ac{hoi} and physical commonsense---in perceiving occlusions~\cite{termine1987perceptual,kellman1983perception}, tracking small objects even if contained by other objects~\cite{feigenson2003tracking}, realizing object permanence~\cite{baillargeon1985object}, recognizing rational \ac{hoi}~\cite{woodward1999infants,skerry2013first}, understanding intuitive physic~\cite{gergely2002developmental,needham1997factors,baillargeon2004infants}, and using exploratory play to understand the environment~\cite{stahl2015observing}. All the evidence calls for a treatment to integrate \ac{hoi} and physical commonsense with a modern computer vision system for scene understanding.

In contrast, few attempts have been made to achieve this goal. This challenge is difficult partially due to the fact that the algorithm has to \emph{jointly} accomplish both 3D holistic scene understanding task and the 3D human pose estimation task in a \emph{physically plausible} fashion. Since this task is beyond the scope of holistic scene understanding in the literature, we define this comprehensive task as \emph{holistic$^{++}$ scene understanding}---to simultaneously estimate human pose, objects, room layout, and camera pose, all in 3D.

Based on one single-view image, existing work either focuses only on 3D holistic scene understanding~\cite{huang2018holistic,zou2017complete,bansal2016marr,song2017semantic} or 3D human pose estimation~\cite{zhao2017simple,ramakrishna2012reconstructing,fang2018learning}. Although one can achieve an impressive performance in a single task by training with an enormous amount of annotated data, we, however, argue that these two tasks are intertwined tightly since the indoor scenes are invented and constructed by human designs to support the daily activities, generating affordance for rich tasks and human activities~\cite{gibson1979ecological}.

To solve the proposed \emph{holistic$^{++}$ scene understanding} task, we attempt to address four fundamental challenges:
\begin{enumerate}[leftmargin=*,noitemsep,nolistsep]
    \item How to utilize the coupled nature of human pose estimation and holistic scene understanding, and make them benefit each other? How to reconstruct the scene with complex human activities and interactions?
    \item How to constrain the solution space of the 3D estimations from a single 2D image?
    \item How to make a physically plausible and stable estimation for complex scenes with human agents and objects?
    \item How to improve the generalization ability to achieve a more robust reconstruction across different datasets? 
\end{enumerate}

To address the first two challenges, we take a novel step to incorporate \textbf{\ac{hoi}} as constraints for \textbf{joint parsing} of both 3D human pose and 3D scene. The integration of \ac{hoi} is inspired by crucial observations of human 3D scene perception, which are challenging for existing systems. Take \autoref{fig:pg} as an example; humans are able to impose a constraint and infer the relative position and orientation between the girl and chair by recognizing the girl is sitting in the chair. Similarly, such a constraint can help to recover the small objects (\eg, recognizing keyboard by detecting the girl is using a computer in \autoref{fig:pg}). By learning \ac{hoi} priors and using the inferred \ac{hoi} as visual cues to adjust the fine-grained spatial relations between human and scene (objects and room layout), the geometric ambiguity (3D estimation solution space) in the single-view reconstruction would be largely eased, and the reconstruction performances of both tasks would be improved.

To address the third challenge, we incorporate \textbf{physical commonsense} into the proposed method. Specifically, the proposed method reasons about the physical relations (\eg, support relation) and penalizes the physical violations to predict a physically plausible and stable 3D scene. The \ac{hoi} and physical commonsense serve as \textbf{general prior} knowledge across different datasets, thus help address the fourth issue. 

To jointly parse 3D human pose and 3D scene, we represent the configuration of an indoor scene by a parse graph shown in \autoref{fig:pg}, which consists of a parse tree with hierarchical structure and a \ac{mrf} over the terminal nodes, capturing the rich contextual relations among human, objects, and room layout. The optimal parse graph to reconstruct both the 3D scene and human poses is achieved by a \ac{map} estimation, where the prior characterizes the prior distribution of the contextual \ac{hoi} and physical relations among the nodes. The likelihood measures the similarity between (i) the detection results directly from 2D object and pose detector, and (ii) the 2D results projected from the 3D parsing results. The parse graph can be iteratively optimized by sampling an \ac{mcmc} with simulated annealing based on posterior probability. The joint optimization relies less on a specific training dataset since it benefits from the prior of \ac{hoi} and physical commonsense which are almost invariant across environments and datasets, and other knowledge learned from well-defined vision task (\eg, 3D pose estimation, scene reconstruction), improving the generalization ability significantly across different datasets compared with purely data-driven methods. 

Experimental results on PiGraphs~\cite{savva2016PiGraphs}, Watch-n-Patch~\cite{wu2015watch}, and SUN RGB-D~\cite{song2015sun} demonstrate that the proposed method outperforms state-of-the-art methods for both 3D scene reconstruction and 3D pose estimation. Moreover, the ablative analysis shows that the \ac{hoi} prior improves the reconstruction, and the physical common sense helps to make physically plausible predictions. 

This paper makes four major contributions:
\begin{enumerate}[leftmargin=*,noitemsep,nolistsep]
    \item We propose a new \emph{holistic$^{++}$ scene understanding} task with a computational framework to jointly infer human poses, objects, room layout, and camera pose, all in 3D.
    
    \item We integrate \ac{hoi} to bridge the human pose estimation and the scene reconstruction, reducing geometric ambiguities (solution space) of the single-view reconstruction.
    
    \item We incorporate physical commonsense, which helps to predict physically plausible scenes and improve the 3D localization of both humans and objects.
    
    \item We demonstrate the joint inference improves the performance of each sub-module and achieves better generalization ability across various indoor scene datasets compared with purely data-driven methods.
\end{enumerate}

\subsection{Related Work}

\paragraph{Single-view 3D Human Pose Estimation:}
Previous methods on 3D pose estimation can be divided into two streams: (i) directly learning 3D pose from a 2D image~\cite{simo2012single,li20143d}, and (ii) cascaded frameworks that first perform 2D pose estimation and then reconstruct 3D pose from the estimated 2D joints~\cite{zhao2017simple,mehta2017vnect,ramakrishna2012reconstructing,wu2016single,cho2016complex,tome2017lifting}. Although these researches have produced impressive results in scenarios with relatively clean background, the problem of estimating the 3D pose in a typical indoor scene with arbitrary cluttered objects has rarely been discussed. Recently, Zanfir \etal~\cite{zanfir2018monocular} adopts constraints of ground plane support and volume occupancy by multiple people, but the detailed relations between human and scene (objects and layout) are still missing. In contrast, the proposed model not only estimates the 3D poses of multiple people with an absolute scale but also models the physical relations between humans and 3D scenes.

\setstretch{0.96}

\paragraph{Single-view 3D Scene Reconstruction:}
Single-view 3D scene reconstruction has three main approaches: (i) Predict room layouts by extracting geometric features to rank 3D cuboids proposals~\cite{zou2017complete,song2017semantic,izadinia2017im2cad,zou2018layoutnet}. (ii) Align object proposals to RGB or depth image by treating objects as geometric primitives or CAD models~\cite{bansal2016marr,song2014sliding,zhou2014learning}. (iii) Joint estimation of the room layout and 3D objects with contexts~\cite{song2017semantic,zhao2013scene,choi2013understanding,zhang2017physically,zou2017complete}. A more recent work by Huang \etal~\cite{huang2018holistic} models the hierarchical structure, latent human context, physical constraints, and jointly optimizes in an analysis-by-synthesis fashion; although human context and functionality were taken into account, indoor scene reconstruction with human poses and \ac{hoi} remains untouched.

\paragraph{Human-Object Interaction:}
Reasoning fine-grained human interactions with objects is essential for a more holistic indoor scene understanding as it provides crucial cues for human activities and physical interactions. In robotics and computer vision, prior work has exploited human-object relations in event, object, and scene modeling, but most work focuses on human-object relation detection in images~\cite{chao2018learning,qi2018learning,mallya2016learning,kjellstrom2011visual}, probabilistic modeling from multiple data sources~\cite{wei2013modeling,savva2014scenegrok,gupta2009observing}, and snapshots generation or scene synthesis~\cite{savva2016PiGraphs, ma2016action,qi2018human,jiang2018configurable}. Different from all previous work, we use the learned 3D \ac{hoi} priors to refine the relative spatial relations between human and scene, enabling a top-down prediction of interacted objects.

\paragraph{Physical Commonsense:}
The ability to infer hidden physical properties is a well-established human cognitive ability~\cite{mccloskey1983intuitive,kubricht2017intuitive}. By exploiting the underlying physical properties of scenes and objects, recent efforts have demonstrated the capability of estimating both current and future dynamics of static scenes~\cite{wu2015galileo,mottaghi2016newtonian} and objects~\cite{zhu2015understanding}, understanding the support relationships and stability of objects~\cite{zheng2013beyond}, volumetric and occlusion reasoning~\cite{silberman2012indoor,zheng2015scene}, inferring the hidden force~\cite{zhu2016inferring}, and reconstructing the 3D scene~\cite{huang2018cooperative,du2018learning} and 3D pose~\cite{zanfir2018monocular}. In addition to the physical properties and support relations among objects adopted in previous methods, we further model the physical relations (i) between human and objects, and (ii) between human and room layout, resulting in a physically plausible and stable scene.

\section{Representation}\label{sec:representation}

The configuration of an indoor scene is represented by a parse graph $pg = (pt, E)$; see \autoref{fig:pg}. It combines a parse tree $pt$ and contextual relations $E$ among the leaf nodes. Here, a parse tree $pt = (V, R)$ includes the vertex set with a three-level hierarchical structure $V = V_r \cup V_m \cup V_t$ and the decomposing rules $R$, where the root node $V_r$ represents the overall scene, the middle node $V_m$ has three types of nodes (objects, human, and room layout), and the terminal nodes $V_t$ contains child nodes of the middle nodes, representing the detected instances of the parent node in this scene. $E \subset V_t \times V_t$ is the set of contextual relations among the terminal nodes, represented by horizontal links.

\textbf{Terminal Nodes {$\mathbf{V_t}$}} in $pg$ can be further decomposed as $V_t = V_{\text{layout}} \cup V_{\text{object}} \cup V_{\text{human}}$. Specifically:
\setstretch{0.97}
\begin{itemize}[leftmargin=*,noitemsep,nolistsep]
    \item The room layout $v \in V_{\text{layout}}$ is represented by a 3D bounding box $X^{L}\in \mathbb{R}^{3 \times 8}$ in the world coordinate. The 3D bounding box is parametrized by the node's attributes, including its 3D size $S^{L}\in \mathbb{R}^{3}$, center $C^{L}\in \mathbb{R}^{3}$, and orientation $Rot(\theta^{L})\in \mathbb{R}^{3\times3}$. See the supplementary for the parametrization of the 3D bounding box.
    
    \item Each 3D object $v \in V_{\text{object}}$ is represented by a 3D bounding box with its semantic label. We use the same 3D bounding box parameterization as the one for the room layout.
    
    \item Each human $v \in V_{\text{human}}$ is represented by 17 3D joints $X^{H}\in \mathbb{R}^{3\times17}$ with their action labels. These 3D joints are parametrized by the pose scale $S^{H}\in \mathbb{R}$, pose center $C^{H}\in \mathbb{R}^{3}$ (\ie, hip), local joint position $Rel^{H}\in \mathbb{R}^{3\times17}$, and pose orientation $Rot(\theta^{H})\in \mathbb{R}^{3\times3}$. Each person is also attributed by a concurrent action label $a$, which is a multi-hot vector representing the current actions of this person: one can ``sit'' and ``drink'', or ``walk'' and ``make phone call'' at the same time.
\end{itemize}

\textbf{Contextual Relations $\mathbf{E}$} contains three types of relations in the scene $E = \{E_s, E_c, E_{hoi}\}$. Specifically:
\begin{itemize}[leftmargin=*,noitemsep,nolistsep]
    \item $E_s$ and $E_c$ denote support relation and physical collision, respectively. These two relations penalize the physical violations among objects, between objects and layout, and between human and layout, resulting in a physically plausible and stable prediction. 
    \item $E_{hoi}$ models \ac{hoi} and provides strong and fine-grained constraints for holistic scene understanding. For instance, if a person is detected as sitting on a chair, we can constrain the relative 3D positions between this person and chair using a pre-learned spatial relation of ``sitting.''
\end{itemize}

\section{Probabilistic Formulation}

The parse graph $pg$ is a comprehensive interpretation of the observed image $I$~\cite{zhu2007stochastic}. The goal of the holistic$^{++}$ scene understanding is to infer the optimal parse graph $pg^*$ given $I$ by an \ac{map} estimation:
\begin{equation}\small
\begin{aligned}
    pg^{*} &= \underset{pg}{\argmax} \, p(pg | I) = \underset{pg}{\argmax} \, p(pg) \cdot p(I | pg)\\
    &= \underset{pg}{\argmax} \, \frac{1}{Z}\exp\{-\mathcal{E}_{phy}(pg)-\mathcal{E}_{hoi}(pg)-\mathcal{E}(I | pg)\}.
\label{eq:equ_pg_prob}
\end{aligned}
\end{equation}
We model the joint distribution by a Gibbs distribution, where the prior probability of parse graph can be decomposed into physical prior $\mathcal{E}_{phy}(pg)$ and \ac{hoi} prior $\mathcal{E}_{hoi}(pg)$; balancing factors are neglected for simplicity.

\textbf{Physical Prior} $\mathcal{E}_{phy}(pg)$ represents physical commonsense in a 3D scene. We consider two types of physical relations among the terminal nodes: support relation $E_s$ and collision relation $E_c$. Therefore, the energy of physical prior is defined as $\mathcal{E}_{phy}(pg) = \mathcal{E}_{s}(pg) + \mathcal{E}_{c}(pg)$. Specifically:

\noindent{\small \textbullet} \emph{Support Relation} $\mathcal{E}_{s}(pg)$ defines the energy between the supported object/human and the supporting object/layout:
\begin{equation}\small
    \mathcal{E}_{s}(pg) = \sum_{(v_i,v_j)\in E_s} \mathcal{E}_{o}(v_i,v_j) + \mathcal{E}_{\text{height}}(v_i,v_j),
    \label{eq:equ_pg_energy}
\end{equation}
where $\mathcal{E}_{o}(v_i,v_j) = 1 - \text{area}(v_i \cap v_j)/\text{area}(v_i)$ is the overlapping ratio in the xy-plane, and $\mathcal{E}_{\text{height}}(v_i,v_j)$ is the absolute height difference between the lower surface of the supported object $v_i$ and the upper surface of the supporting object $v_j$; $\mathcal{E}_{o}(v_i,v_j) = 0$ when the supporting object is the floor and $\mathcal{E}_{\text{height}}(v_i,v_j) = 0$ when the supporting object is the wall.

\noindent{\small \textbullet} \emph{Physical Collision} $\mathcal{E}_{c}(pg)$ denotes the physical violations. We penalize the intersection among human, objects, and room layout except the objects in \ac{hoi} and objects that could be a container. The potential function is defined as:
\begin{equation}
\resizebox{.9\hsize}{!}{$\displaystyle
    \mathcal{E}_{c}(pg) = \mathop{\sum \mathcal{C}(v, V_{\text{layout}})}_{v \in (V_{\text{object}} \cup V_{\text{human}})} +
    \mathop{\sum \mathcal{C}(v_{i}, v_{j})}_{\substack{v_{i} \in V_{\text{object}}\\ v_{j} \in V_{\text{human}}\\ (v_{i}, v_{j}) \notin E_{hoi}} } +
    \mathop{\sum \mathcal{C}(v_{i}, v_{j})}_{\substack{v_i, v_j \in V_{\text{object}}\\ v_i, v_j \notin V_{\text{container}}}},
$}%
\end{equation}
where $\mathcal{C}()$ denotes the volume of intersection between entities. $V_{\text{container}}$ denotes the objects that can be a container, such as a cabinet, desk, and drawer. 

\textbf{Human-object Interaction Prior} $\mathcal{E}_{hoi}(pg)$ is defined by the interactions between human and objects:
\begin{equation}\small
    \mathcal{E}_{hoi}(pg) = \sum_{(v_i, v_j)\in E_{hoi}} \mathcal{K}(v_i,v_j,a_{v_j}),
    \label{eq:hoi}
\end{equation}
where $v_i \in V_{\text{object}},v_j \in V_{\text{human}}$, and $\mathcal{K}$ is an \ac{hoi} function that evaluates the interaction between an object and a human given the action label $a$:
\begin{equation}\small
    \mathcal{K}(v_i,v_j,a_{v_j}) = -\log l(v_i,v_j  | a_{v_j}),
\end{equation}
where $l(v_i,v_j  | a_{v_j})$ is the likelihood of the relative position between node $v_i$ and $v_j$ given an action label $a$. We formulate the action detection as a \emph{multi-label classification}; see \autoref{sec:action_detection} for details. The likelihood $l(\cdot)$ models the distance between key joints and the center of the object; \eg, for ``sitting,'' it models the relative spatial relation between the hip and the center of a chair. The likelihood can be learned from 3D \ac{hoi} datasets with a multivariate Gaussian distribution $(\Delta x, \Delta y, \Delta z) \sim \mathcal{N}_{3}(\boldmath{\mu}, \boldmath{\Sigma})$, where $\Delta x, \Delta y$, and $\Delta z$ are the relative distances in the directions of three axes.

\textbf{Likelihood $\mathcal{E}(I | pg)$} characterizes the consistency between the observed 2D image and the inferred 3D result. The projected 2D object bounding boxes and human poses can be computed by projecting the inferred 3D objects and human poses onto a 2D image plane. The likelihood is obtained by comparing the directly detected 2D bounding boxes and human poses with projected ones from inferred 3D results:
\begin{equation}
\resizebox{.9\hsize}{!}{$\displaystyle
    \mathcal{E}(I | pg) = \mathop{\sum}_{v \in V_{\text{object}}} \cdot \mathcal{D}_{o}(B(v), B^{\prime}(v)) + \mathop{\sum}_{v \in V_{\text{human}}} \cdot \mathcal{D}_{h}(Po(v), Po^{\prime}(v)),
$}%
\end{equation}
where $B()$ and $B^{\prime}()$ are the bounding boxes of detected and projected 2D objects, $Po()$ and $Po^{\prime}()$ the poses of detected and projected 2D humans, $\mathcal{D}_{o}(\cdot)$ the \ac{iou} between the detected 2D bounding box and the convex hull of the projected 3D bounding box, and $\mathcal{D}_{h}(\cdot)$ the average pixel-wise Euclidean distance between two 2D poses.

\setstretch{0.99}

\begin{figure}[t!]
    \centering
    \includegraphics[width=0.95\linewidth,trim={0cm 0.5cm 0cm 0.1cm},clip]{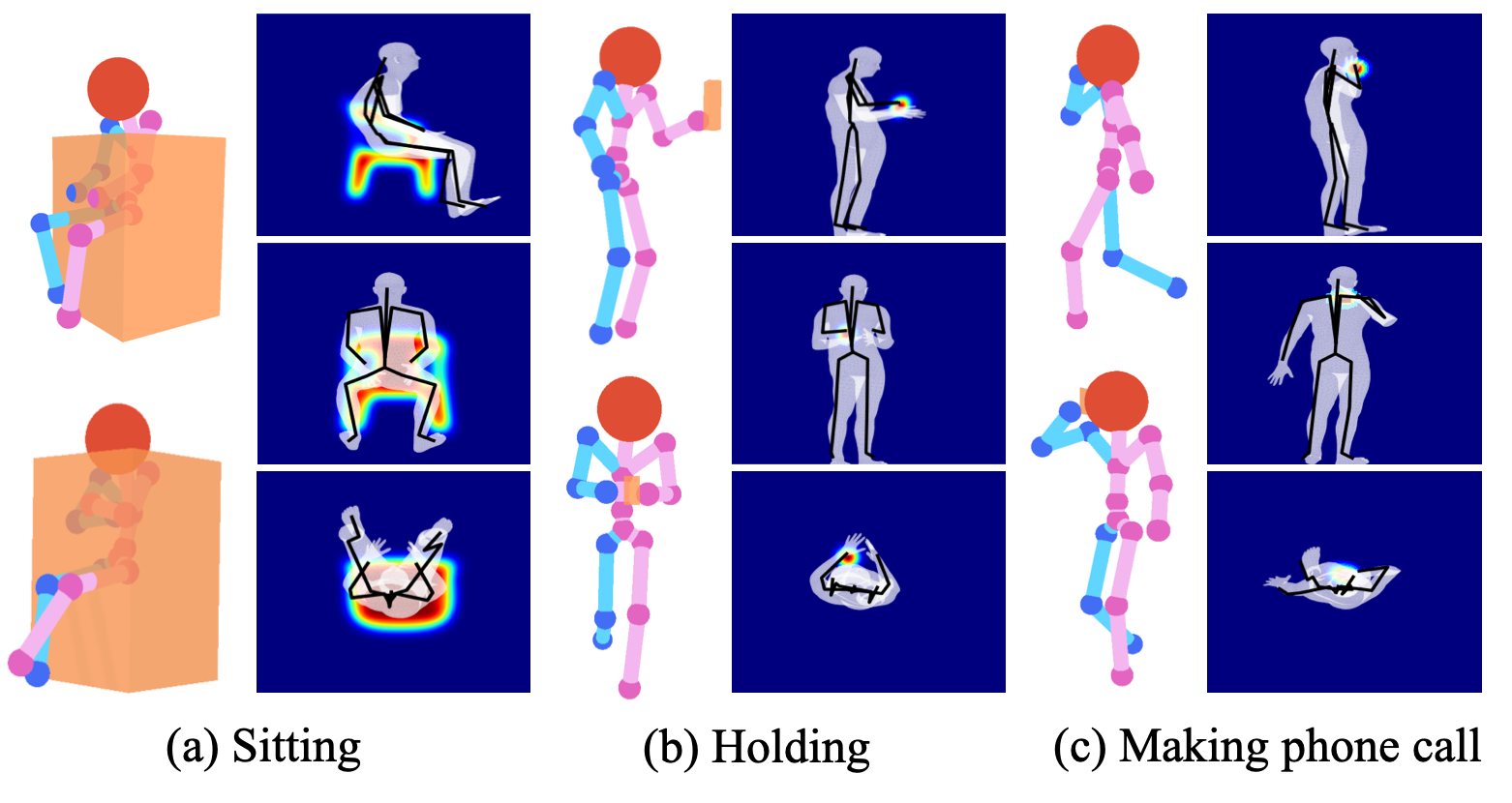}
    \caption{Examples of typical \ac{hoi}s and examples from the SHADE dataset. The heatmap indicates the probable locations of \ac{hoi}.}
    \label{fig:hoi_dataset}
\end{figure}

\section{SHADE Dataset}\label{sec:shade}

We collect SHADE (Synthetic Human Activities with Dynamic Environment), a self-annotated dataset that consists of dynamic 3D human skeletons and objects, to learn the prior model for each \ac{hoi}. It is collected from a video game Grand Theft Auto V with various daily activities and \ac{hoi}s. Currently, there are over 29 million frames of 3D human poses, where 772,229 frames are annotated. On average, each annotated frame is associated with 2.03 action labels and 0.89 \ac{hoi}s. The SHADE dataset contains 19 fine-grained \ac{hoi}s for both indoor and outdoor activities. By selecting most frequent \ac{hoi}s and merging similar \ac{hoi}s, we choose 6 final HOIs: \emph{read [phone, notebook, tablet], sit-at [human-table relation], sit [human-chair relation], make-phone-call, hold, use-laptop}. \autoref{fig:hoi_dataset} shows some typical examples and relations in the dataset.

\section{Joint Inference}

Given a single RGB image as the input, the goal of joint inference is to find the optimal parse graph that maximizes the posterior probability $p(pg | I)$. The joint parsing is a four-step process:
(i) 3D scene initialization of the camera pose, room layout, and 3D object bounding boxes,
(ii) 3D human pose initialization that estimates rough 3D human poses in a 3D scene, 
(iii) concurrent action detection, and
(iv) joint inference to optimize the objects, layout, and human poses in 3D scenes by maximizing the posterior probability.

\subsection{3D Scene Initialization} 

Following~\cite{huang2018cooperative}, we initialize the 3D objects, room layout, and camera pose cooperatively, where the room layout and objects are parametrized by 3D bounding boxes. For each object $v_i \in V_{\text{object}}$, we find its supporting object/layout by minimizing the supporting energy:
\begin{equation}
\resizebox{.9\hsize}{!}{$\displaystyle
    v_j^{*} = \underset{v_j}{\argmin}\, \mathcal{E}_{o}(v_i,v_j) + \mathcal{E}_{\text{height}}(v_i,v_j)-\lambda_{s}\log\,p_{spt}(v_i,v_j),
$}%
\end{equation}
where $v_j \in (V_{\text{object}}, V_{\text{layout}})$ and $p_{spt}(v_i,v_j)$ are the prior probabilities of the supporting relation modeled by multinoulli distributions, and $\lambda_{s}$ a balancing constant. 

\setstretch{0.96}

\subsection{3D Human Pose Initialization}

We take 2D poses as the input and predict 3D poses in a local 3D coordinate following~\cite{tome2017lifting}, where the 2D poses are detected and estimated by~\cite{cao2017realtime}. The local 3D coordinate is centered at the human hip joint, and the z-axis is aligned with the up direction of the world coordinate.

To transform this local 3D pose into the world coordinate, we find the 3D world coordinate $\mathbf{v_{3D}}\in \mathbb{R}^{3}$ of one visible 2D joint $\mathbf{v_{2D}}\in \mathbb{R}^{2}$ (\eg, head) by solving a linear equation with the camera intrinsic parameter $K$ and estimated camera pose $R$. Per the pinhole camera projection model, we have
\begin{equation}
    \alpha
    \begin{bmatrix}
    \mathbf{v_{2D}}\\
    1
    \end{bmatrix}
    = K \cdot R \cdot \mathbf{v_{3D}},
\end{equation}
where $\alpha$ is a scaling factor in the homogeneous coordinate. To make the function solvable, we assume a pre-defined height $h_0$ for the joint position $\mathbf{v_{3D}}$ in the world coordinate. Lastly, the 3D pose initialization is obtained by aligning the local 3D pose and the corresponding joint position with $\mathbf{v_{3D}}$.

\subsection{Concurrent Action Detection}\label{sec:action_detection}

We formulate the concurrent action detection as a multi-label classification problem to ease the ambiguity in describing the action. We define a portion of the action labels (\eg, ``eating'', ``making phone call'') as the \ac{hoi} labels, and the remaining action labels (\eg, ``standing'', ``bending'') as general human poses without \ac{hoi}. The mixture of \ac{hoi} actions and non-\ac{hoi} actions covers most of the daily human actions in indoor scenes. We manually map each of the \ac{hoi} action labels to a 3D \ac{hoi} relation learned from the SHADE dataset, and use the \ac{hoi} actions as cues to improve the accuracy of 3D reconstruction by integrating it as prior knowledge in our model. The concurrent action detector takes 2D skeletons as the input and predicts multiple action labels with a three-layer \ac{mlp}.

The dataset for training the concurrent action detectors consists of both synthetic data and real-world data. It is collected from:
(i) The synthetic dataset described in \autoref{sec:shade}. We project the 3D human poses of different \ac{hoi}s into 2D poses with random camera poses. 
(ii) The dataset proposed and collected by~\cite{joo2017panoptic}, which also contains 3D poses of multiple persons in social interactions. We project 3D poses into 2D following the same method as in (i).
(iii) The 2D poses in an action recognition dataset~\cite{yao2011human}. Our results show that the synthetic data can significantly expand the training set and help to avoid overfitting in concurrent action detection.

\begin{algorithm}[t!]
\caption{Joint Inference Algorithm}\label{alg:joint_inf}
\begin{algorithmic}
    \State \textbf{Given}: Image $I$, initialized parse graph $pg_{init}$
    \Procedure{Phase 1}{}
        \For {Different temperatures}
            \State \multiline{%
            Inference with physical commonsense $\mathcal{E}_{phy}$ but without \ac{hoi} $\mathcal{E}_{hoi}$: randomly select from room layout, objects, and human poses to optimize $pg$}%
        \EndFor
    \EndProcedure
    \Procedure{Phase 2}{}
        \State Match each agent with their interacting objects
    \EndProcedure
    \Procedure{Phase 3}{}
        \For {Different temperatures}
            \State \multiline{%
            Inference with total energy $\mathcal{E}$, including physical commonsense and \ac{hoi}: randomly select from layout, objects, and human poses to optimize $pg$}%
        \EndFor
    \EndProcedure
    \Procedure{Phase 4}{}
        \State Top-down sampling by \ac{hoi}s
    \EndProcedure
\end{algorithmic}
\end{algorithm}

\subsection{Inference}

Given an initialized parse graph, we use \ac{mcmc} with simulated annealing to jointly optimize the room layout, 3D objects, and 3D human poses through the non-differentiable energy space; see \autoref{alg:joint_inf} as a summary. To improve the efficiency of the optimization process, we adopt a scheduling strategy that divides the optimization process into following four phases with different focuses:
(i) Optimize objects, room layout, and human poses without \ac{hoi}s.
(ii) Assign \ac{hoi} labels to each agent in the scene, and search the interacting objects of each agent.
(iii) Optimize objects, room layout, and human poses jointly with \ac{hoi}s.
(iv) Generate possible miss-detected objects by top-down sampling. 

\paragraph{Dynamics:} In Phase (i) and (iii), we use distinct \ac{mcmc} processes. To traverse non-differentiable energy spaces, we design Markov chain dynamics ${q_{1}^{o}, q_{2}^{o}, q_{3}^{o}}$ for objects, ${q_{1}^{l}, q_{2}^{l}}$ for room layout, and ${q_{1}^{h}, q_{2}^{h}, q_{3}^{h}}$ for human poses.

${\bullet}$ Object Dynamics: Dynamics $q_{1}^{o}$ adjusts the position of an object, which translates the object center in one of the three Cartesian coordinate axes or along the depth direction; the depth direction starts from the camera position and points to the object center. Translation along depth is effective with proper camera pose initialization. Dynamics $q_{2}^{o}$ proposes rotation of the object with a specified angle. Dynamics $q_{3}^{o}$ changes the scale of the object by expanding or shrinking corner positions of the cuboid with respect to the object center. Each dynamic can diffuse in two directions: translate in the direction of `$+x$' and `$-x$,' or rotate in the direction of clockwise and counterclockwise. To better traverse in energy space, the dynamics may propose to move along the gradient descent direction with a probability of 0.95 or the gradient ascent direction with a probability of 0.05.

${\bullet}$ Human Dynamics: Dynamics $q_{1}^{h}$ proposes to translate 3D human joints along x, y, z, or depth direction. Dynamics $q_{2}^{h}$ rotates the human pose with a certain angle. Dynamics $q_{3}^{h}$ adjusts the scale of human poses by a scaling factor on the 3D joints with respect to the pose center.

${\bullet}$ Layout Dynamics: Dynamics $q_{1}^{l}$ translates the wall towards or away from the layout center. Dynamics $q_{2}^{l}$ adjusts the floor height, equivalent to changing the camera height.

\begin{figure*}[t!]
    \centering
    \includegraphics[width=\linewidth,trim={0.6cm 3.3cm 0.6cm 4cm},clip]{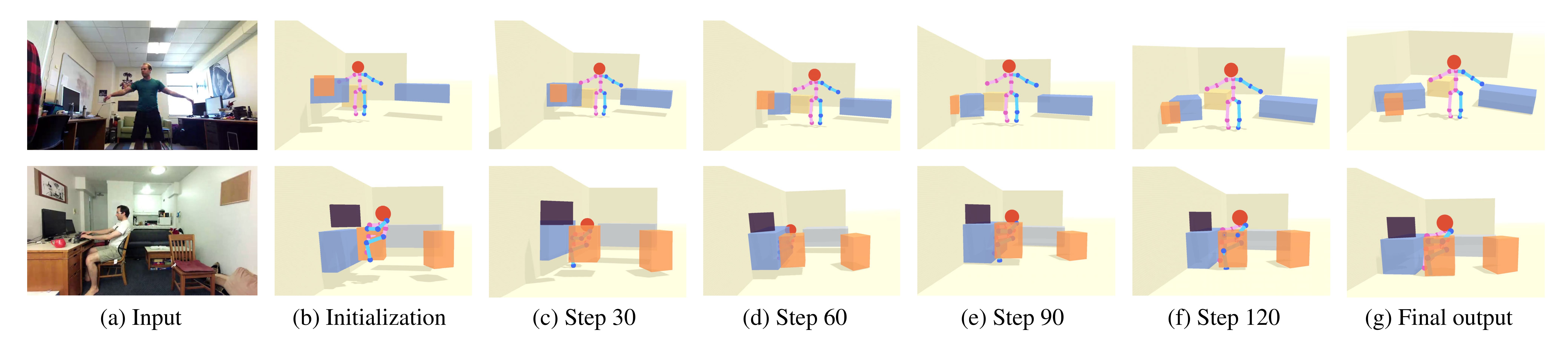}
    \caption{The optimization process of the scene configuration by simulated annealing \ac{mcmc}. Each step is the number of accepted proposal.}
    \label{fig:opt_process}
\end{figure*}

\begin{figure}[t!]
    \centering
    \includegraphics[width=\linewidth,trim={0cm 2.7cm 0cm 0cm},clip]{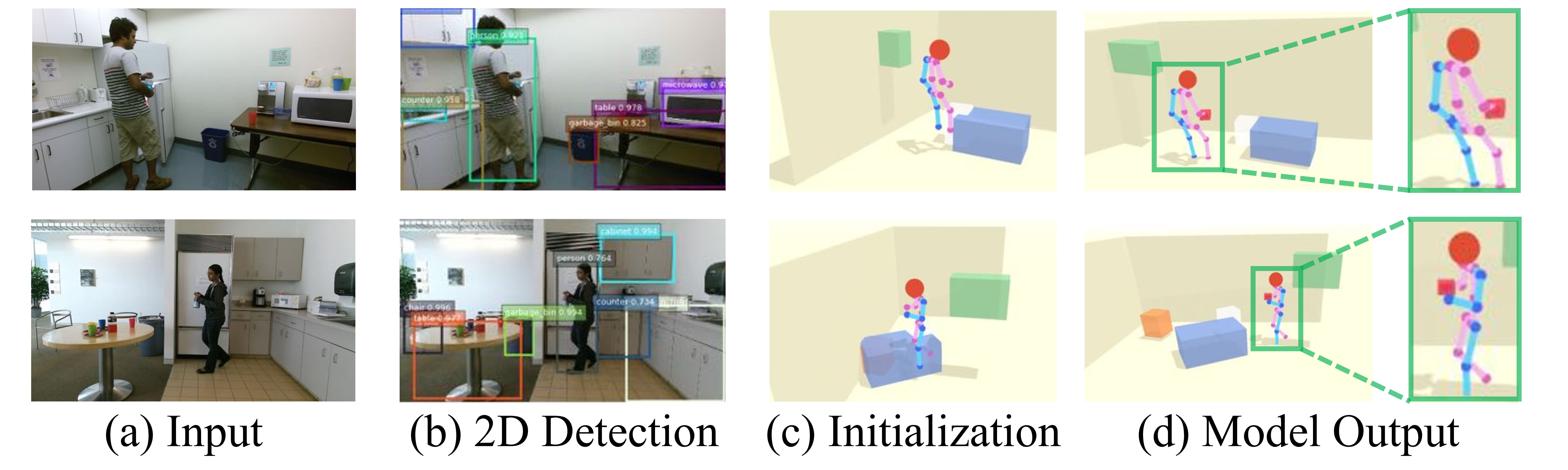}
    \caption{Illustration of the top-down sampling process. The object detection module misses the detection of the bottle held by the person, but our model can still recover the bottle by reasoning \ac{hoi}.}
    \label{fig:missobj}
\end{figure}

In each sampling iteration, the algorithm proposes a new $pg^{\prime}$ from current $pg$ under the proposal probability of $q(pg\to pg^{\prime} | I)$ by applying one of the above dynamics. The generated proposal is accepted with respect to an acceptance rate $\alpha(\cdot)$ as in the Metropolis-Hastings algorithm~\cite{hastings1970monte}:
\begin{equation}
    \alpha(pg\to pg^{\prime}) = \min(1, \frac{q(pg^{\prime}\to pg)\cdot p(pg^{\prime}   | I)}{q(pg\to pg^{\prime})\cdot p(pg | I)}),
\end{equation}%
\setstretch{0.97}%
A simulated annealing scheme is adopted to obtain $pg$ with a high probability.

\textbf{Top-down sampling:} By top-down sampling objects from \ac{hoi}s relations, the proposed method can recover the interacting 3D objects that are too small or novel to be detected by the state-of-the-art 2D object detector. In Phase (iv), we propose to sample an interacting object from the person if the confidence of \ac{hoi} is higher than a threshold; we minimize the \ac{hoi} energy in \autoref{eq:hoi} to determine the category and location of the object; see examples in \autoref{fig:missobj}.

\textbf{Implementation Details:}
In Phase (ii), we search the interacting objects for each agent involved in \ac{hoi} by minimizing the energy in \autoref{eq:hoi}.
In Phase (iii), after matching each agent with their interacting objects, we can jointly optimize objects, room layout, and human poses with the constraint imposed by \ac{hoi}. \autoref{fig:opt_process} shows examples of the simulated annealing optimization process.

\section{Experiments}\label{sec:exp}

Since the proposed task is new and challenging, limited data and state-of-the-art methods are available for the proposed problem. For fair evaluations and comparisons, we evaluate the proposed algorithm on three types of datasets: (i) Real data with full annotation on PiGraphs dataset~\cite{savva2016PiGraphs} with limited 3D scenes. (ii) Real data with partial annotation on daily activity dataset Watch-n-Patch~\cite{wu2015watch}, which only contains ground-truth depth information and annotations of 3D human poses. (iii) Synthetic data with generated annotations to serve as the ground truth: we sample 3D human poses of various activities in SUN RGB-D dataset~\cite{song2015sun} and project the sampled skeletons back onto the 2D image plane.

\setstretch{0.965}

\subsection{Comparative methods}

To the best of our knowledge, no previous algorithm jointly optimizes the 3D scene and 3D human pose from a single image. Therefore, we compare our model against state-of-the-art methods for each task. Particularly, we compare with~\cite{huang2018cooperative} for single-image 3D scene reconstruction and VNect~\cite{mehta2017vnect} for 3D pose estimation in the world coordinate.

Since VNect can only estimate a single person, we design an additional baseline for 3D multi-person human pose estimation in the world coordinate. We first extract a 2048-D image feature vector using the Global Geometry Network (GGN)~\cite{huang2018cooperative} to capture the global geometry of the scene. The concatenated vector (GGN image feature, 2D pose, 3D pose in the local coordinate, and the camera intrinsic matrix) is fed into a 5-layer fully connected network to predict the 3D pose. The fully-connected layers are trained using the mean squared error loss. We train the network on the training set of the synthetic SUN RGB-D dataset. Please refer to supplementary materials for more details of the baseline model.

\subsection{Dataset}

\textbf{PiGraphs}~\cite{savva2016PiGraphs} contains 30 scenes and 63 video recordings obtained by Kinect v2, designed to associate human poses with object arrangements. There are 298 actions available in approximately 2-hours of recordings. Each recording is about 2-minute long, with an average 4.9 action annotation. We removed the frames with no human appearance or annotations, resulting in 36,551 test images.

\textbf{Watch-n-Patch} (WnP)~\cite{wu2015watch} is an activity video dataset recorded by Kinect v2. It contains several human daily activities as compositions of multiple actions interacting with various objects. The dataset comes with activity annotations, depth maps, and 3D human poses. We test our algorithm on 1,210 randomly selected frames.

\textbf{SUN RGB-D}~\cite{song2015sun} contains rich indoor scenes that are densely annotated with 3D bounding boxes, room layouts, and camera poses. The original dataset has 5,050 testing images, but we discarded images with no detected 2D objects, invalid 3D room layout annotation, limited space, or small field of view, resulting in 3,476 testing images.

\begin{figure}[t!]
    \centering
    \includegraphics[width=\linewidth,trim={0cm 1cm 0cm 0.3cm},clip]{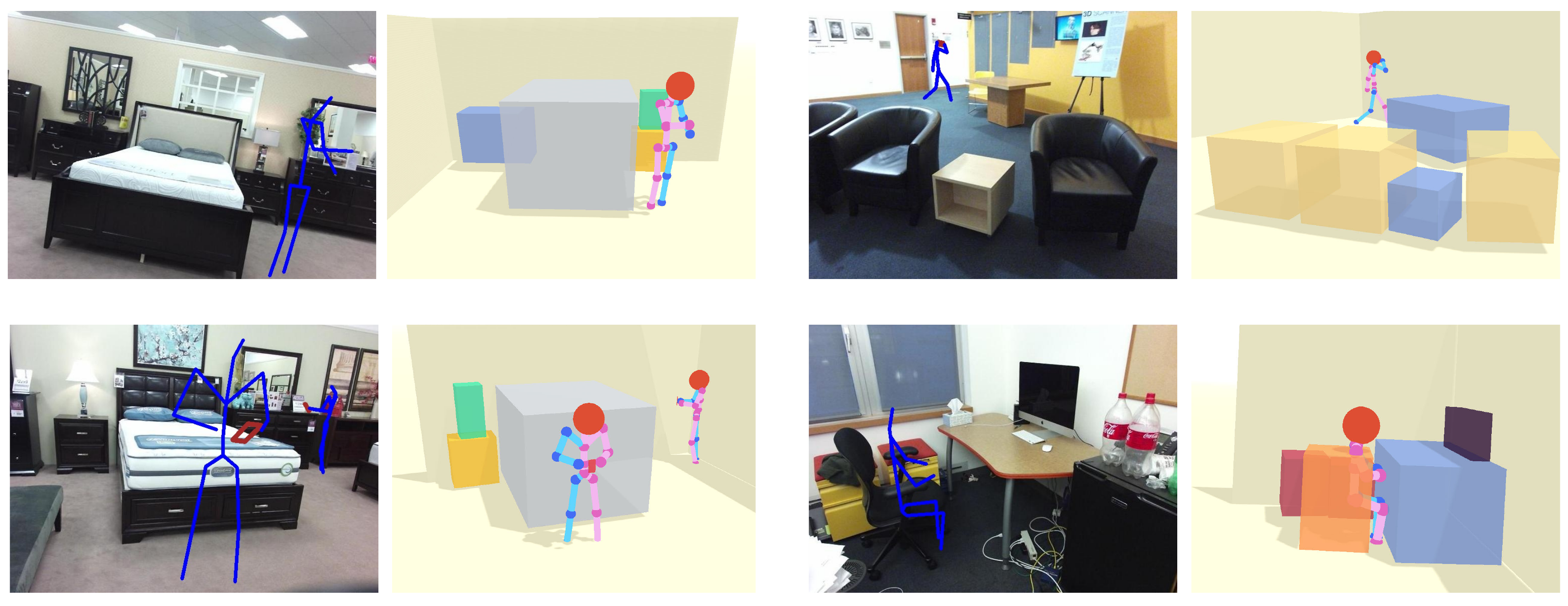}
    \caption{Augmenting SUN RGB-D with synthetic human poses.}
    \label{fig:syn_runrgbd}
\end{figure}

\textbf{Synthetic SUN RGB-D} is augmented from SUN RGB-D dataset by sampling human poses in the scenes. Following methods of sampling imaginary human poses in~\cite{huang2018holistic}, we extend the sampling to more generalized settings for various poses. The augmented human is represented by a 6-tuple $\langle a, \mu, t, r, s, \hat{\mu} \rangle$, where $a$ is the action type, $\mu$ the pose template, $t$ translation, $r$ rotation, $s$ scale, and $\hat{\mu} = \mu\cdot r\cdot s + t$ the imagined human skeleton. For each action label, we sample an imagined human pose inside a 3D scene: $\langle t^{*}, r^{*}, s^{*} \rangle = \underset{t,r,s}{\argmin} \, \mathcal{E}_{phy} + \mathcal{E}_{hoi}$. If $a$ is involved with any \ac{hoi} unit, we further augment the 3D bounding box of the object. After sampling a human pose, we project the augmented 3D scenes back onto the 2D image plane using the ground truth camera matrix and camera pose; see examples in \autoref{fig:syn_runrgbd}. For a fair comparison of 3D human pose estimation on synthetic SUN RGB-D, all the algorithms are provided with the ground truth 2D skeletons as the input.

For 3D scene reconstruction, both~\cite{huang2018cooperative} and the proposed 3D scene initialization are learned using SUN RGB-D training data and tested on the above three datasets. For 3D pose estimation, both~\cite{mehta2017vnect} and the initialization of the proposed method are trained on public datasets, while the baseline is trained on synthetic SUN RGB-D. Note that we only use the SHADE dataset for learning a dictionary of \ac{hoi}s.

\subsection{Quantitative and Qualitative Results}

We evaluate the proposed model on holistic$^{++}$ scene understanding task by comparing the performances on both 3D scene reconstruction and 3D pose estimation.

\begin{figure*}[t!]
    \centering
    \includegraphics[width=0.96\linewidth,trim={0cm 1cm 0cm 0.5cm},clip]{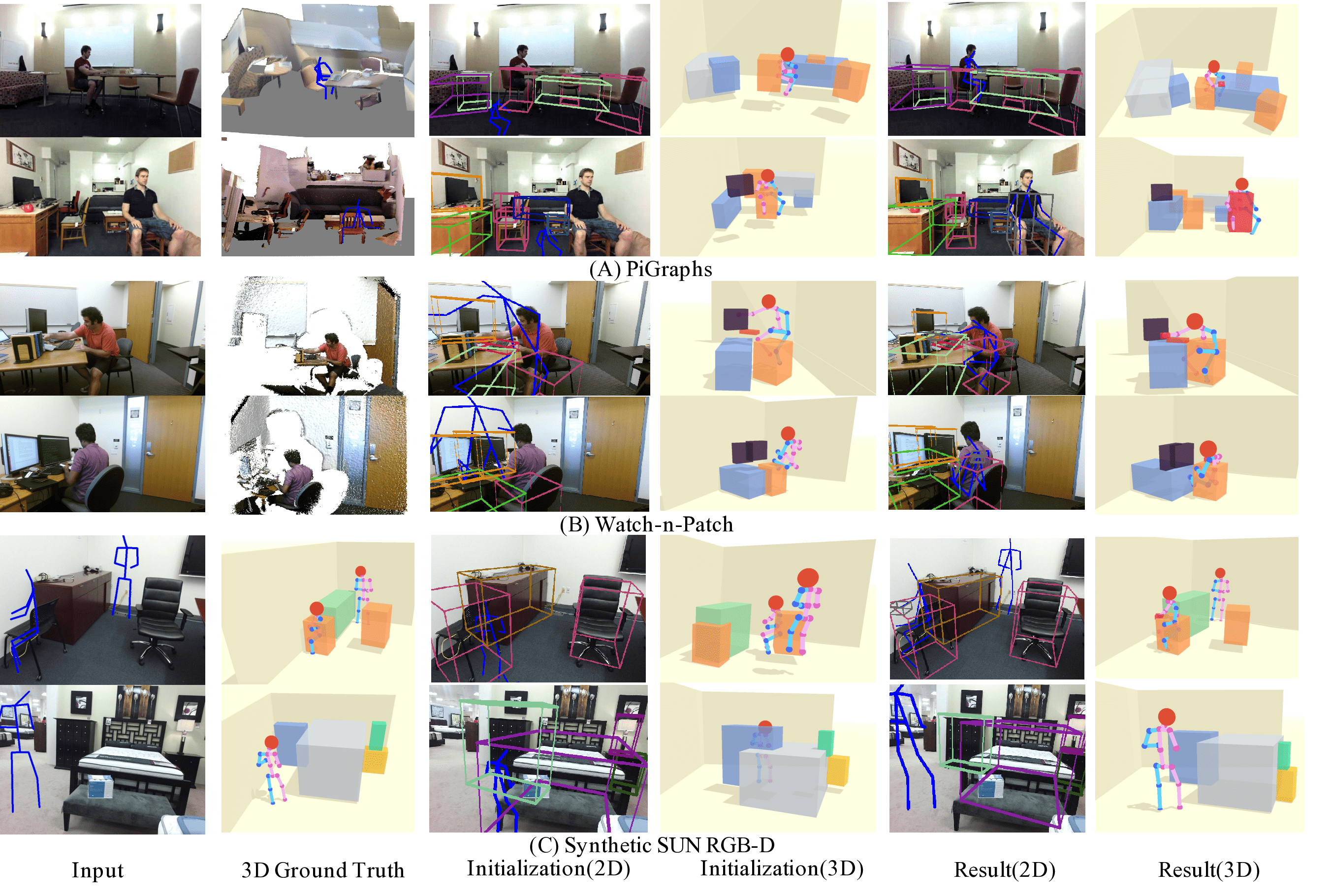}
    \caption{Qualitative results of the proposed method on three datasets. The proposed model improves the initialization with accurate spatial relations and physical plausibility and demonstrates an outstanding generalization across various datasets.}
    \label{fig:res_pose}
\end{figure*}

\paragraph{Scene Reconstruction:} We compute the 3D \ac{iou} and 2D \ac{iou} of object bounding boxes to evaluate the 3D scene reconstruction and the consistency between the 3D world and 2D image. Following the metrics described in~\cite{huang2018cooperative}, we compute the 3D \ac{iou} between the estimated 3D bounding boxes and the annotated 3D bounding boxes on PiGraphs and SUN RGB-D. For dataset without ground-truth 3D bounding boxes (\ie, Watch-n-Patch), we evaluate the distance between the camera center and the 3D object center. To evaluate the 2D-3D consistency, the 2D \ac{iou} is computed between the projected 2D boxes of the 3D object bounding boxes and the ground-truth 2D boxes or detected 2D boxes (\ie, Watch-n-Patch). As shown in \autoref{tab:res_obj}, the proposed method improves the state-of-the-art 3D scene reconstruction results on all three datasets without specific training on each of them. More importantly, it significantly improves the results on PiGraphs and Watch-n-Patch compared with~\cite{huang2018cooperative}. The most likely reason is:~\cite{huang2018cooperative} is trained on SUN RGB-D dataset in a purely data-driven fashion, therefore difficult to generalize across to other datasets (\ie, PiGraphs, and Watch-n-Patch). In contrast, the proposed model incorporates more general prior knowledge of \ac{hoi} and physical commonsense, and combines such knowledge with 2D-3D consistency (likelihood) for joint inference, avoiding the over-fitting caused by the direct 3D estimation from 2D. \autoref{fig:res_pose} shows the qualitative results on all three datasets.

\begin{table}[t!]
    \centering
    \caption{Quantitative Results of 3D Scene Reconstruction}
    \resizebox{1.0\linewidth}{!}{%
    \setlength\tabcolsep{2pt}%
    \begin{tabular}{c||c|c|c||c|c|c}
    \hline
        \rowcolor{mygray}
         Methods & \multicolumn{3}{c||}{Huang \etal \cite{huang2018cooperative}} & \multicolumn{3}{c}{Ours}\\ \hline
         \rowcolor{mygray}
         Metric & 2D IoU (\%) & 3D IoU (\%) & Depth (m) & 2D IOU (\%) & 3D IoU (\%) & Depth (m)\\ \hline
     PiGraphs & 68.6 & 21.4 & - & \bf 75.1 & \bf 24.9 & - \\ 
         SUN RGB-D & 63.9 & 17.7 & - & \bf 72.9 & \bf 18.2 & - \\ 
         WnP & 67.3 & - & 0.375 & \bf 73.6 & - & \bf 0.162 \\ \hline
    \end{tabular}}
    \label{tab:res_obj}
\end{table}

\paragraph{Pose Estimation:} We evaluate the pose estimation in both 3D and 2D. For 3D evaluation, we compute the Euclidean distance between the estimated 3D joints and the 3D ground-truth and average it over all the joints. For 2D evaluation, we project the estimated 3D pose back to the 2D image plane and compute the pixel distance against the ground truth. See \autoref{tab:res_pose} for quantitative results. The proposed method outperforms two other methods in both 2D and 3D. On the synthetic SUN RGB-D dataset, all algorithms are given the ground truth 2D poses as the input for a fair comparison. Although the baseline model achieves better performances since the baseline model fits well for the 3D human poses synthesized with limited templates, the 3D poses estimated by VNect and baseline model deviate a lot from the ground truth for datasets with real human poses (\ie, PiGraph, and Watch-n-Patch). In contrast, the proposed algorithm performs consistently well, demonstrating an outstanding generalization ability across various datasets.

\begin{table}[t!]
    \centering
    \caption{Quantitative Results of Global 3D Pose Estimation}
    \small
    \resizebox{0.95\linewidth}{!}{%
    \setlength\tabcolsep{2pt}%
    \begin{tabular}{c||c|c||c|c||c|c}
    \hline
        \rowcolor{mygray}
        Methods & \multicolumn{2}{c||}{VNect\cite{mehta2017vnect}} & \multicolumn{2}{c||}{Baseline} & \multicolumn{2}{c}{Ours}\\ \hline
         \rowcolor{mygray}
         Metrics & 2D (pix) & 3D (m) & 2D (pix) & 3D (m) & 2D (pix) & 3D (m)\\ \hline
         PiGraphs & 63.9 & 0.732 & 284.5 & 2.67 & \bf 15.9 & \bf 0.472 \\ 
         SUNRGBD & - & - & 45.81 & \bf 0.435 & \bf 14.03 & 0.517 \\ 
         WnP & 50.51 & 0.646 & 325.2 & 2.14 & \bf 20.5 & \bf 0.330 \\ \hline
    \end{tabular}}
\label{tab:res_pose}
\end{table}

\begin{table}[t!]
    \centering
    \caption{Ablative results of \ac{hoi} on 3D object \ac{iou} (\%), 3D pose estimation error (m), and miss-detection rate (MR, \%)}
    \small
    \resizebox{0.9\linewidth}{!}{%
    \setlength\tabcolsep{2pt}%
    \begin{tabular}{c||c|c|c||c|c|c}
    \hline
        \rowcolor{mygray}
        Methods& \multicolumn{3}{c||}{\emph{w/o hoi}} & \multicolumn{3}{c}{\emph{Full model}}\\ \hline
        \rowcolor{mygray}
        HOI Type  & Object $\uparrow$ & Pose $\downarrow$ & MR $\downarrow$ & Object $\uparrow$ & Pose $\downarrow$ & MR  $\downarrow$\\ \hline \hline
         Sit & 26.9 & 0.590 & 15.2 & \bf 27.8 & \bf 0.521 & \bf 13.1 \\ \hline
         Hold & 17.4 & 0.517 & 78.9 & \bf 17.6 & \bf 0.490 & \bf 54.6 \\ \hline
         Use Laptop & 14.1 & 0.544 & 58.8 & \bf 15.0 & \bf 0.534 & \bf 43.3 \\ \hline
         Read & \bf 14.5 & 0.466 & 65.3 & 14.3 & \bf 0.453 & \bf 41.9 \\ \hline
    \end{tabular}}
\label{tab:abl_hoi}
\end{table}

\begin{figure}[t!]
    \centering
    \includegraphics[width=\linewidth,trim={0cm 0.2cm 0cm 0cm},clip]{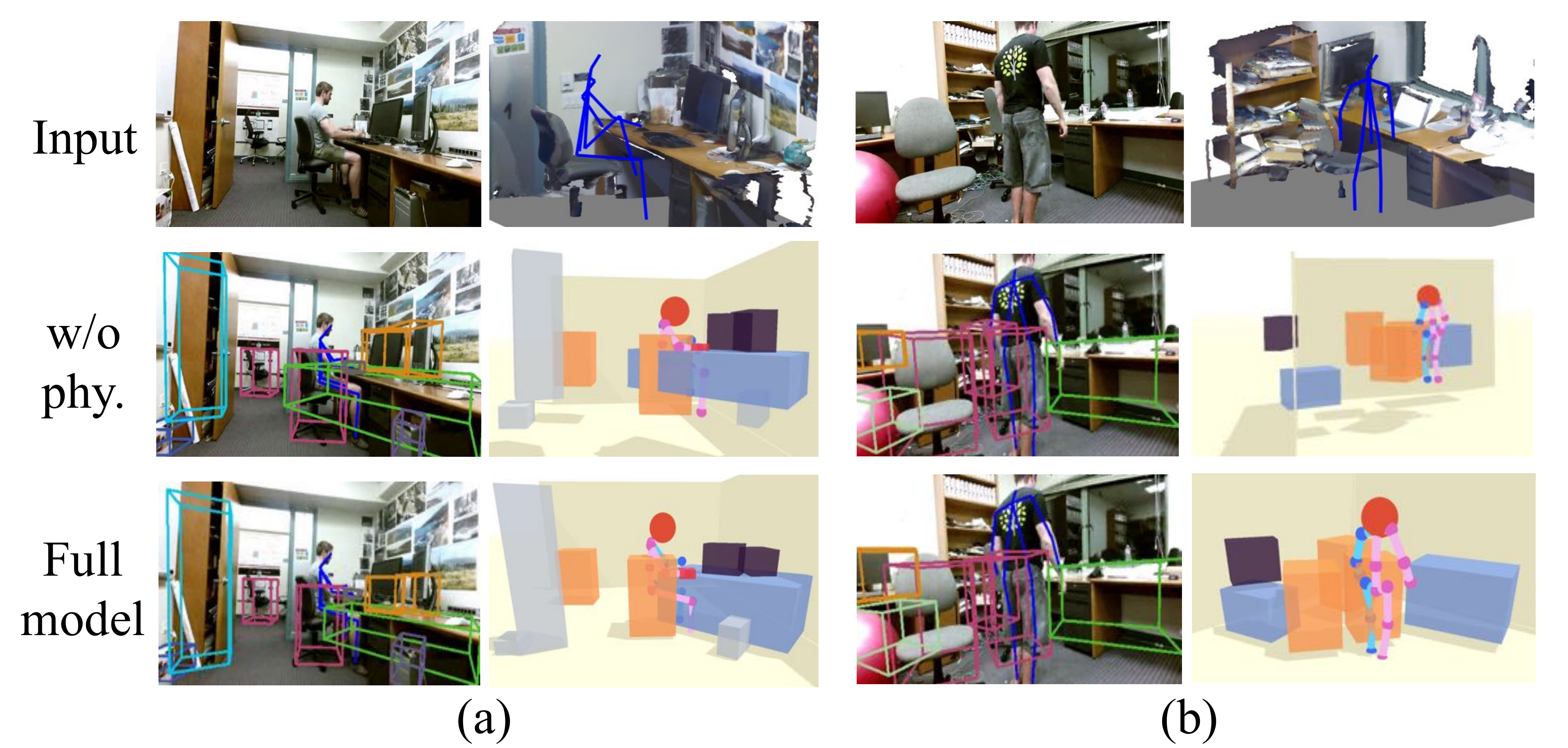}
    \caption{Qualitative comparison between (a) model \emph{w/o phy.} and (b) the full model on PiGraphs dataset.}
    \label{fig:abl_physical}
\end{figure}

\paragraph{Ablative Analysis:} To analyze the contributions of \ac{hoi} and physical commonsense, we compare two variants of the proposed full model: (i) model \emph{w/o \ac{hoi}}: without \ac{hoi} $\mathcal{E}_{hoi}(pg)$, and (ii) model \emph{w/o phy.}: without physical commonsense $\mathcal{E}_{phy}(pg)$. 

\noindent{\small \textbullet} \emph{Human-Object Interaction}. We compare our full model with model \emph{w/o hoi} to evaluate the effects of each category of \ac{hoi}. Evaluation metrics include 3D pose estimation error, 3D bounding box \ac{iou}, and miss-detection rate (MR) of the objects interacted with agents. The experiments are conducted on PiGraphs dataset and Synthetic SUN RGB-D dataset with the annotated \ac{hoi} labels. Note that for the consistency of the ablative analysis across three different datasets, we merge the \emph{sit} and \emph{sit-at} into \emph{sit}, and eliminate the \emph{make-phone-call}. As shown in \autoref{tab:abl_hoi}, the performances of both scene reconstruction and human pose estimation are hindered without reasoning \ac{hoi}, indicating \ac{hoi} helps to infer the relative spatial relationship between agents and objects to improve the performance of both two tasks further. Moreover, a marked performance gain of miss-detection rate implies the effectiveness of the top-down sampling process during the joint inference.

\noindent{\small \textbullet} \emph{Physical Commonsense}. Reasoning about physical commonsense drives the reconstructed 3D scene to be physically plausible and stable. We test 3D estimation of object bounding boxes on the PiGraphs dataset using \emph{w/o phy.} and the full model. The full model outperforms \emph{w/o phy.} in two aspects: (i) 3D object detection \ac{iou} (from 23.5\% to 24.9\%), and (ii) physical violation (from 0.223m to 0.150m); see qualitative comparisons in \autoref{fig:abl_physical}. The physical violation is computed as the distance between the lower surface of an object and the upper surface of its supporting object. Objects detected by model \emph{w/o phy.} may float in the air or penetrate each other, while the full model yields physically plausible results.

\vspace{-6pt}
\section{Conclusion}
\vspace{-3pt}

This paper tackles a challenging holistic$^{++}$ scene understanding problem to jointly solve 3D scene reconstruction and 3D human pose estimation from a single RGB image. By incorporating physical commonsense and reasoning about \ac{hoi}, our approach leverages the coupled nature of these two tasks and goes beyond merely reconstructing the 3D scene or human pose by reasoning about the concurrent action of human in the scene. We design a joint inference algorithm which traverses the non-differentiable solution space with \ac{mcmc} and optimizes the scene configuration. Experiments on PiGraphs, Watch-n-Patch, and Synthetic SUN RGB-D demonstrate the efficacy of the proposed algorithm and the general prior knowledge of \ac{hoi} and physical commonsense.

\noindent\textbf{Acknowledgments:}
We thank Tengyu Liu from UCLA CS department for providing the SHADE dataset. The work reported herein was supported by DARPA XAI grant N66001-17-2-4029, ONR MURI grant N00014-16-1-2007, and ONR robotics grant N00014-19-1-2153.

\clearpage
\setstretch{0.94}
{\small
\bibliographystyle{ieee_fullname}
\bibliography{egbib}
}

\end{document}